\documentclass{article} 
\usepackage[preprint]{colm2025_conference}
\usepackage{microtype}
\usepackage{hyperref}
\usepackage{url}
\usepackage{booktabs}
\usepackage{enumitem}
\usepackage{caption}
\usepackage{graphicx} 
\usepackage{tikz}
\usepackage{pgfplots}
\usepackage{amsmath}
\usepackage{subcaption}
\usepackage{listings}
\usepackage{xcolor}
\pgfplotsset{compat=1.18}

\usepackage{lineno}

\definecolor{darkblue}{rgb}{0, 0, 0.5}

\hypersetup{colorlinks=true, citecolor=darkblue, linkcolor=darkblue, urlcolor=darkblue}

\title{ScholarCopilot: Training Large Language Models for Academic Writing with Accurate Citations}

\author{Yubo Wang\textsuperscript{1,$\dagger$}, Xueguang Ma\textsuperscript{1,$\dagger$}, Ping Nie\textsuperscript{3}, Huaye Zeng\textsuperscript{1}, Zhiheng Lyu\textsuperscript{1}, Yuxuan Zhang\textsuperscript{1}, \\[0pt]
\textbf{Benjamin Schneider\textsuperscript{1}, Yi Lu\textsuperscript{1}, Xiang Yue\textsuperscript{2}, Wenhu Chen\textsuperscript{1,4,$\dagger$}} \\
\textsuperscript{1}University of Waterloo,
\textsuperscript{2}Carnegie Mellon University, Pittsburgh,\\
\textsuperscript{3}Independent Researcher,
\textsuperscript{4}Vector Institute, Toronto\\
}

\usepackage{footnote}

\begin{document}
\footnotetext[0]{$\dagger$ Core Contributors}
\footnotetext[1]{$\dagger$ Project website: \url{https://tiger-ai-lab.github.io/ScholarCopilot/}}
\ifcolmsubmission
\linenumbers
\fi

\maketitle


\vspace{-1em}
\begin{abstract}
\vspace{-0.5em}
Academic writing requires both coherent text generation and precise citation of relevant literature. Although recent Retrieval-Augmented Generation (RAG) systems have significantly improved factual accuracy in general-purpose text generation, their ability to support professional academic writing remains limited. In this work, we introduce ScholarCopilot, a unified framework designed to enhance existing large language models for generating professional academic articles with accurate and contextually relevant citations.
ScholarCopilot dynamically determines when to retrieve scholarly references by generating a retrieval token [RET], which is then used to query a citation database. The retrieved references are fed into the model to augment the generation process.
We jointly optimize both the generation and citation tasks within a single framework to improve efficiency. Our model is built upon Qwen-2.5-7B and trained on 500K papers from arXiv. It achieves a top-1 retrieval accuracy of 40.1\% on our evaluation dataset, outperforming baselines such as E5-Mistral-7B-Instruct (15.0\%) and BM25 (9.8\%).
On a dataset of 1,000 academic writing samples, ScholarCopilot scores 16.2/25 in generation quality—measured across relevance, coherence, academic rigor, completeness, and innovation—significantly surpassing all existing models, including much larger ones like the Retrieval-Augmented Qwen2.5-72B-Instruct. Human studies further demonstrate that ScholarCopilot, despite being a 7B model, significantly outperforms ChatGPT, achieving 100\% preference in citation quality and over 70\% in overall usefulness.

\end{abstract}

\begin{figure}[h]
\centering
\includegraphics[width=0.82\linewidth]{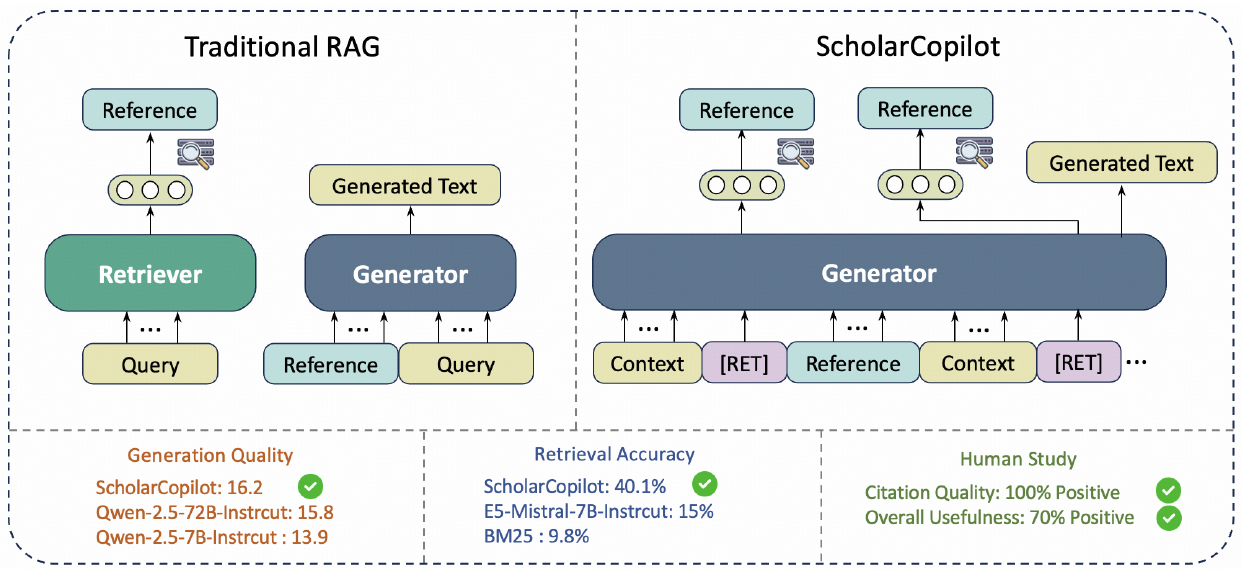}
\caption{Comparison of traditional Retrieval-Augmented Generation (RAG) systems and our proposed \textbf{ScholarCopilot}. Traditional RAG systems (left) separately perform retrieval and generation, leading to representation misalignment. In contrast, ScholarCopilot (right) dynamically generates retrieval tokens (\texttt{[RET]}) during text generation for integrated and context-aware reference retrieval.}
\vspace{-1.2em}
\label{fig:teaser}
\end{figure}

\section{Introduction}

Academic writing is a knowledge-intensive task that requires both structured content generation and accurate citation of relevant literature. While large language models (LLMs) such as GPT-4~\citep{gpt4}, Deepseek-v3~\citep{deepseekv3}, and Qwen2.5~\citep{qwen2} can generate fluent academic-style text, they frequently hallucinate citations, undermining their reliability for research writing~\citep{hallucination_survey_1,hallucination_survey_2}.

\begin{figure}[ht]
\centering
\includegraphics[width=\linewidth]{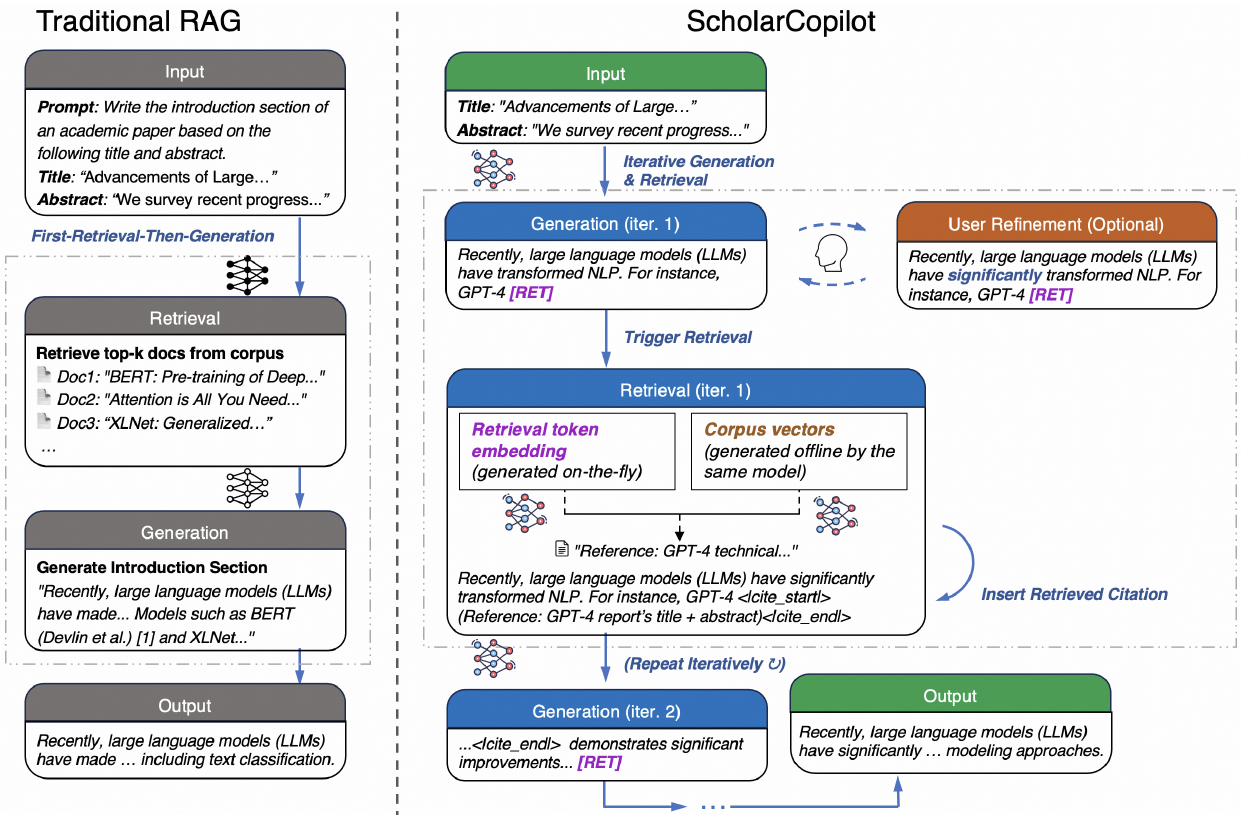}
\caption{Comparison between traditional Retrieval-Augmented Generation (RAG) methods (left) and ScholarCopilot (right). Traditional RAG follows a static retrieval-then-generation pipeline, retrieving references independently before generation. ScholarCopilot dynamically interleaves retrieval and generation by producing retrieval tokens (\texttt{[RET]}) based on current context, enabling context-aware citation retrieval and optional user refinement.}
\label{fig:intro_figure}
\end{figure}

Recent advanced retrieval-augmented generation (RAG)~\citep{lewisrag, shi2023replugretrievalaugmentedblackboxlanguage} systems
address this issue by retrieving relevant references from external knowledge bases to enhance factual consistency and reduce hallucinations.
These approaches typically follow a \textbf{first-retrieve-then-generate} pipeline, as illustrated in Figure~\ref{fig:intro_figure} (left), where retrieval is conducted independently prior to generation.
However, this pipeline neglects the evolving generation context, making it difficult to dynamically adjust retrieval decisions based on the changing information needs during writing. For instance, when generating an introduction mentioning GPT-4, traditional approaches cannot adaptively retrieve GPT-4-related references precisely when needed, since retrieval decisions are predetermined without awareness of the specific generation context.
Consequently, these methods suffer from three key limitations: (1) separate optimization of retrieval and generation models leads to misalignment in query intent;
(2) predetermined retrieval decisions lack flexibility and context-awareness;
(3) static pipeline limits user control over the generation of content and citation needs.

To overcome these limitations, we propose \textbf{ScholarCopilot}, an agentic RAG framework tailored for assisting academic paper writing that seamlessly integrates text generation and citation retrieval in a unified, iterative manner, as illustrated in Figure~\ref{fig:intro_figure} (right).
Instead of relying on separate retrieval and generation stages, ScholarCopilot dynamically determines when retrieval is necessary by generating special \textbf{retrieval tokens (\texttt{[RET]})} based on the evolving generation context.
Upon generating these tokens, ScholarCopilot pauses the generation process, retrieves relevant scholarly references, and integrates their content (abstracts or key excepts) directly back into subsequent generation steps.
The dense representations of these retrieval tokens are optimized via contrastive learning, enabling efficient similarity search.
Additionally, ScholarCopilot allows optional user refinement and citation triggering during the iterative process, providing flexibility to integrate human domain expertise for further improving generation quality.
This unified, iterative approach enhances citation accuracy, improves content coherence, and maintains efficiency without additional overhead.

We evaluate ScholarCopilot extensively on academic writing tasks, focusing on \textbf{generation quality, retrieval accuracy, and overall user experience}. Our model achieves \textbf{40.1\% top-1 retrieval accuracy}, significantly surpassing baselines such as \textbf{E5-Mistral-7B-Instruct (15.0\%)}~\citep{e5mistral} and \textbf{BM25 (9.8\%)}~\citep{bm25}, with consistent performance gains across all Top-K thresholds.
In terms of generation quality, ScholarCopilot scores \textbf{16.2/25} on a 1000 samples dataset with LLM-as-judge across five dimensions (relevance, coherence, academic rigor, completeness, and innovation), substantially outperforming larger models such as \textbf{Qwen-2.5-7B-Instruct (13.9)} and \textbf{Qwen-2.5-72B-Instruct (15.8)}.
A comprehensive user study with 10 experienced academic writers further confirms ScholarCopilot's effectiveness, particularly highlighting its citation accuracy (100\% positive ratings) and overall usefulness (70\% positive ratings) compared to ChatGPT.

Our main contributions are summarized as follows:
\begin{itemize}[
    itemsep=0pt,        
    parsep=0pt,         
    topsep=0pt,         
    leftmargin=1em      
]
    \item  \textbf{A unified generation-retrieval model} that effectively integrates retrieval into the generative process, enabling seamless citation retrieval while reducing inference overhead and improving citation accuracy and relevance.
    \item \textbf{A comprehensive evaluation framework} that assesses both retrieval accuracy and academic text quality along five critical dimensions: content relevance, logical coherence, academic rigor, information completeness, and scholarly innovation.
    \item \textbf{A large-scale training dataset} consisting of 500k computer science papers from arXiv with comprehensive citation networks (33 matched citations per paper on average), facilitating robust learning of academic writing patterns and citation-aware scholarly practices.
\end{itemize}

\section{Related Work}

\subsection{Dense Retrieval}
Recent studies have demonstrated that dense retrieval methods using pretrained language models to encode text into dense vectors outperform traditional lexical retrievers like TF-IDF and BM25. Following the introduction of DPR~\citep{karpukhin-etal-2020-dense}, several approaches have been proposed to improve dense retrieval through advanced training strategies (e.g., ANCE~\citep{xiong2021approximate}, Condenser~\citep{gao-callan-2021-condenser}), data augmentation techniques (e.g., BGE~\citep{bge_embedding}, GTE~\citep{li2023generaltextembeddingsmultistage}, DRAMA~\citep{drama}), and by leveraging large language models as backbones (e.g., LLM2Vec~\citep{behnamghader2024llmvec}, RepLlama~\citep{repllama}, Mistral-E5~\citep{e5mistral}).
Today, commercial embedding models (e.g., OpenAI~\citep{neelakantan2022textcodeembeddingscontrastive}, GeminiEmbed~\citep{lee2025geminiembeddinggeneralizableembeddings}) are widely used in real-world retrieval systems.
However, most existing methods are designed for single-turn retrieval with short queries for retrieval, which is not well suited for citation suggestions where paper contexts are used to retrieve the next relevant citation.

\subsection{Retrieval Augmented Generation}
In the era of LLM, the Retrieval Augmented Generation (RAG) paradigm integrates a retrieval model for the generation model, allowing the generation model to have access to external knowledge, improving the generation's correctness and factuality for downstream tasks, such as question answering or fact verification~\citep{petroni-etal-2021-kilt}.
Traditional RAG methods typically follow a retrieve-then-generate pipeline~\citep{lewisrag, gao2024retrievalaugmentedgenerationlargelanguage}, where retrieval is conducted independently based on an initial query, and the retrieved documents are concatenated as context for the generation model.
While effective for short-form generation tasks, this static pipeline struggles in scenarios requiring long-form generation with evolving information needs.
To address this limitation, recent methods such as FLARE~\citep{jiang-etal-2023-active} and SelfRAG~\citep{selfrag} propose iterative RAG strategies, where retrieval and generation are interleaved, allowing retrieval decisions to adapt dynamically based on the generation trajectory. These systems demonstrate improved factual accuracy for long-form content by leveraging the generation context to refine retrieval queries.
Recent work, OpenScholar~\citep{openscholar} aims to improve long-form scientific question answering with self-feedback inference in RAG.
However, they still decouple the retrieval and generation models, which can lead to representational misalignment for implicit query intent and increased inference overhead.
More unified approaches, such as GritLM~\citep{muennighoff2025generative} and OneGen~\citep{zhang-etal-2024-onegen}, train a unified model to serve both as the generator and retriever.
These models share representations and can cache hidden states during generation, improving the efficiency of the system.
Despite their advantages, most of these systems have been evaluated primarily on QA-style benchmarks~\citep{popqa} and do not consider the iterative and citation-centric requirements of academic writing.


Our approach differs from prior work in three ways. First, it uses iterative RAG, interleaving retrieval with generation to fit evolving citation needs. Second, it handles implicit intent, inferring citations from context without explicit queries. Third, it enables human-in-the-loop interaction, allowing users to guide or refine citations during writing.

\section{ScholarCopilot}
\subsection{Dataset}
\begin{figure}[t]
\centering
\includegraphics[width=0.98\linewidth]{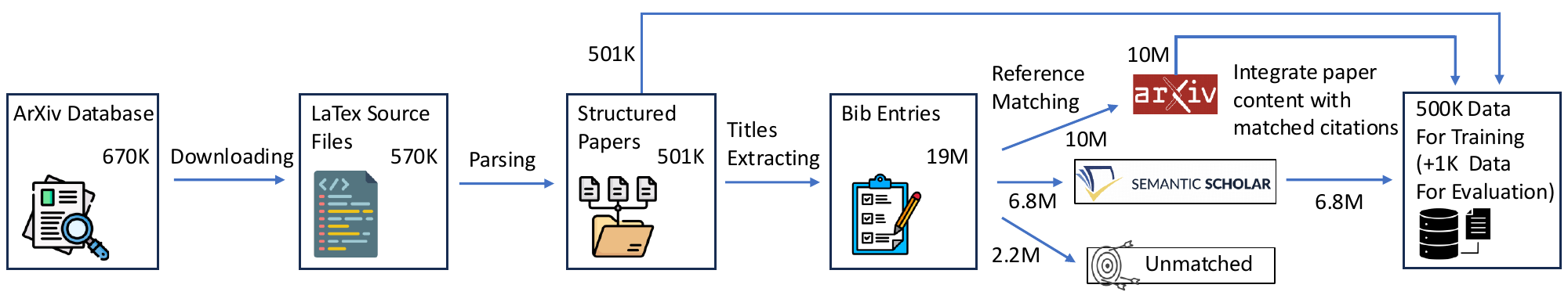}
\caption{The pipeline for creating the ScholarCopilot dataset. Our final dataset includes 10M citations matched from arXiv and 6.8M citations matched from Semantic Scholar (one paper may be cited by multiple articles). However, at inference time, to ensure reference quality, we only use the 670K articles from arXiv as the corpus.}
\label{fig:dataset_creation}
\end{figure}

To train a model capable of accurately generating academic text with appropriate citations, we constructed a large-scale dataset of computer science research papers. Our dataset creation process consisted of five major stages, as illustrated in Figure~\ref{fig:dataset_creation}.

\textbf{Stage 1: Paper Collection.} We collected 670K computer science papers published on arXiv~\citep{arxiv} between 2007 and 2024. From this initial corpus, we successfully obtained LaTeX source code for 570K papers, which formed the foundation of our dataset.

\textbf{Stage 2: Structure Parsing.} We developed heuristic methods to parse the LaTeX source files and extract structured components, including titles, abstracts, introductions, related work sections, and bibliographies. This stage involved handling complex LaTeX formatting and nested environments. After filtering out papers with parsing failures, we retained 501K successfully structured documents (500K for training and 1K for evaluation), preserving the
hierarchical organization essential for understanding academic documents.

\textbf{Stage 3: Citation Extraction.} We extracted citation information from bibliographic entries in each paper. Due to the diversity of BibTeX formatting conventions, regular expression-based approaches proved ineffective for reliable title extraction. Instead, we employed the Qwen-2.5-3B-Instruct~\citep{qwen2} model to robustly extract paper titles from bibliography entries. This approach yielded 19M unique citation titles across our corpus.

\textbf{Stage 4: Reference Matching.} To enable retrieval during training and inference, we matched the extracted citation titles against established academic databases. Of the 19M citation titles, we successfully matched 10M in the arXiv metadata repository and an additional 6.8M in the Semantic Scholar database~\citep{semanticscholar}, resulting in a total of 16.8M matched citations. The remaining unmatched citations typically corresponded to URLs or publications not indexed in either database.

\textbf{Stage 5: Dataset Integration.} Finally, we integrated the parsed paper structures with their matched citations to create the comprehensive ScholarCopilot dataset. The training dataset comprises 500K papers, with 1K papers reserved for evaluation. Each paper contains an average of 38 citations, of which we successfully matched 33 (87\%) to their corresponding entries in academic databases. 

\subsection{Unified Training for Generation and Citation Retrieval}

\begin{figure}[t]
    \centering
    \includegraphics[width=0.95\textwidth]{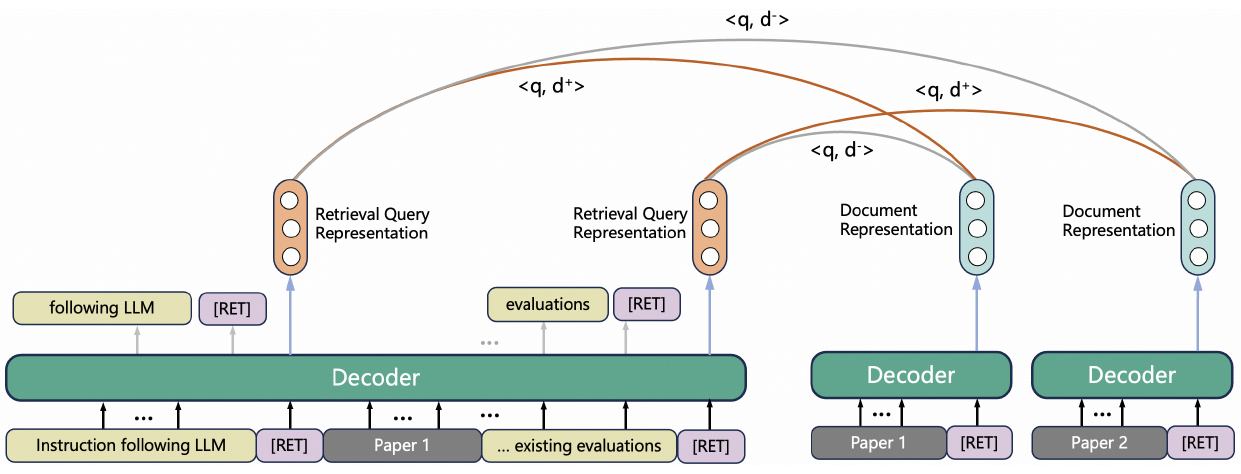}
    \caption{\textbf{Unified training framework of ScholarCopilot.} The architecture jointly optimizes the next token prediction loss for text generation and the contrastive loss for citation retrieval.
    Retrieval tokens ([RET]) dynamically trigger retrieval.
    $<q, d^+>$ indicates the positive pair of query and document during contrastive learning, and $<q, d^->$ indicates the negative pair.
    The generation model and retrieval model share parameters.
    In this figure, Paper 1 and Paper 2 can be considered as hard negatives for each other.
    }
    \label{fig:unified_training}
\end{figure}

ScholarCopilot jointly optimizes two objectives: next token prediction for text generation and contrastive learning for citation retrieval. Figure~\ref{fig:unified_training} illustrates this architecture. Detailed training procedures and hyperparameters can be found in the Appendix~\ref{sec:training-details}.

\textbf{Next Token Prediction Loss \(L_g\).} ScholarCopilot adopts the standard autoregressive language modeling objective for text generation, maximizing the log-likelihood of each token \(x_t\) conditioned on previous tokens \(x_{<t}\) and retrieved content \(c\) (e.g., paper abstracts) when retrieval occurs: \( L_g = -\sum_{t}\log p(x_t|x_{<t}, c). \) Retrieval is dynamically triggered via special tokens (\texttt{[RET]}) generated during inference.\vspace{1ex}\\
\textbf{Contrastive Loss \(L_r\) for Citation Retrieval.} To optimize retrieval token representations, ScholarCopilot employs contrastive learning, encouraging higher similarity between retrieval token embeddings \(q\) and positive (relevant) citation embeddings \(d^+\), and lower similarity with negative (irrelevant) citations \(d^-\). Formally, the contrastive loss is defined as \(L_r = -\log [\exp(\text{sim}(q, d^+)) / (\exp(\text{sim}(q, d^+)) + \sum_{d^-}\exp(\text{sim}(q, d^-)))]\), where \(\text{sim}(\cdot,\cdot)\) denotes cosine similarity. Positive citations are those referenced in the ground-truth paper. Negative citations are obtained through in-batch sampling: citations from the same paper irrelevant to the current context serve as hard negatives, while those from other papers in the batch are easy negatives.\vspace{1ex}\\
\textbf{Joint Optimization.} ScholarCopilot minimizes the combined loss \(L_{total} = L_g + \lambda L_r\), where \(\lambda\) balances generation and retrieval objectives. In our experiments, we set \(\lambda=1\), equally weighting both terms. Joint optimization ensures effective retrieval token learning for accurate citation retrieval without compromising generation quality.

\section{Experiments}

\subsection{Baselines}

We compare ScholarCopilot against several baseline approaches. For generation baselines, we include: \textbf{Qwen-2.5-7B-re}, Qwen-2.5-7B-Instruct enhanced by citation retrieval using E5-Mistral-7B-Instruct; \textbf{Qwen-2.5-72B-re}, the larger 72B parameter variant with the same retrieval method; \textbf{Qwen-2.5-7B-gt}, Qwen-2.5-7B-Instruct provided with ground truth citations as input; and \textbf{Qwen-2.5-72B-gt}, the 72B parameter variant with ground truth citations. Retrieval baselines include \textbf{BM25}~\citep{bm25}, a classical lexical retrieval approach commonly used in information retrieval systems; and \textbf{E5-Mistral-7B-Instruct}~\citep{e5mistral}, a recent embedding-based retrieval model fine-tuned for retrieval tasks.

\subsection{Evaluation Methodology}

We evaluate models across two primary criteria: generation quality and retrieval accuracy.

\textbf{Generation Quality.} We evaluate academic writing quality using five dimensions, each scored from 1 (poor) to 5 (excellent): \textit{Content Relevance} (alignment with academic topic), \textit{Logical Coherence} (clarity and logical flow of arguments), \textit{Academic Rigor} (scholarly depth and precision), \textit{Information Completeness} (coverage comprehensiveness), and \textit{Scholarly Innovation} (originality and insightfulness). 
To ensure reliability, GPT-4o evaluates model-generated content against ground truth texts from 1,000 test set papers across these dimensions. Detailed evaluation prompts are provided in the appendix~\ref{eval_prompt}.\vspace{1ex}\\
\textbf{Retrieval Accuracy.} Citation retrieval is evaluated using Recall@k (k = 1 to 10), defined as the proportion of cases where the correct citation appears among the top-k retrieved results. Specifically, citations and subsequent content in 1,000 test samples are masked, and retrieval models predict citations based solely on the preceding context. For baseline models, we found that using the entire preceding context reduces performance; thus, we only use the last sentence before the citation as the query. Recall@k is computed by comparing predicted citations to the original ground-truth citations.

\subsection{Main Results}
\begin{table}[t]
\centering
\small
\begin{tabular*}{\linewidth}{l|ccccc|c}
\toprule
\textbf{Model} & \textbf{Relevance} & \textbf{Coherence} & \textbf{Academic} & \textbf{Completeness} & \textbf{Innovation} & \textbf{Total} \\
\midrule
\multicolumn{7}{c}{Groundtruth Citations} \\
\midrule
Qwen-2.5-7B-gt & 3.27 & 3.07 & 2.52 & 2.77 & 2.82 & 14.44 \\
Qwen-2.5-72B-gt & 3.73 & 3.71 & 3.00 & 3.11 & 3.28 & 16.82 \\
\midrule
\multicolumn{7}{c}{Retrieved Citations} \\
\midrule
Qwen-2.5-7B-re & 3.16 & 3.30 & 2.26 & 2.41 & 2.80 & 13.94 \\
Qwen-2.5-72B-re & 3.56 & 3.61 & 2.68 & 2.84 & 3.12 & 15.81 \\
\midrule
ScholarCopilot & 3.63 & 3.66 & 2.87 & 2.89 & 3.17 & 16.21 \\
$\Delta$ (Ours - 72B)  & +0.07 &  +0.05 & +0.09 & + 0.04  &  +0.05 & +0.4  \\
\bottomrule
\end{tabular*}
\caption{Generation quality evaluation results by GPT-4o. All scores are on a scale of 1-5, except for Total which is the sum (max 25).}
\vspace{-0.3cm}
\label{tab:main_results}
\end{table}

Table~\ref{tab:main_results} presents the generation quality results for our approach compared to baseline models. ScholarCopilot achieves a total score of 16.21 out of 25, outperforming both Qwen-2.5-7B-Instruct with retrieval enhancement (13.94) and standard Qwen-2.5-7B-Instruct with ground truth citations (14.44). Notably, our approach even surpasses the much larger Qwen-2.5-72B-Instruct model with retrieval enhancement (15.81) and comes close to the 72B model with ground truth citations (16.82), despite having only about 10\% of its parameters.

ScholarCopilot demonstrates particular strengths in Relevance (3.63) and Coherence (3.66), comparable to the 72B models.
The improvement in Academic Rigor (2.87 vs. 2.26 for Qwen-2.5-7B-re) highlights our model's ability to incorporate appropriate citations and scholarly conventions.
These results confirm that our unified approach to generation and citation effectively improves academic writing quality even with a relatively small model.

\subsection{Ablation Studies}

\begin{figure}[ht]
\centering
\includegraphics[width=0.8\linewidth]{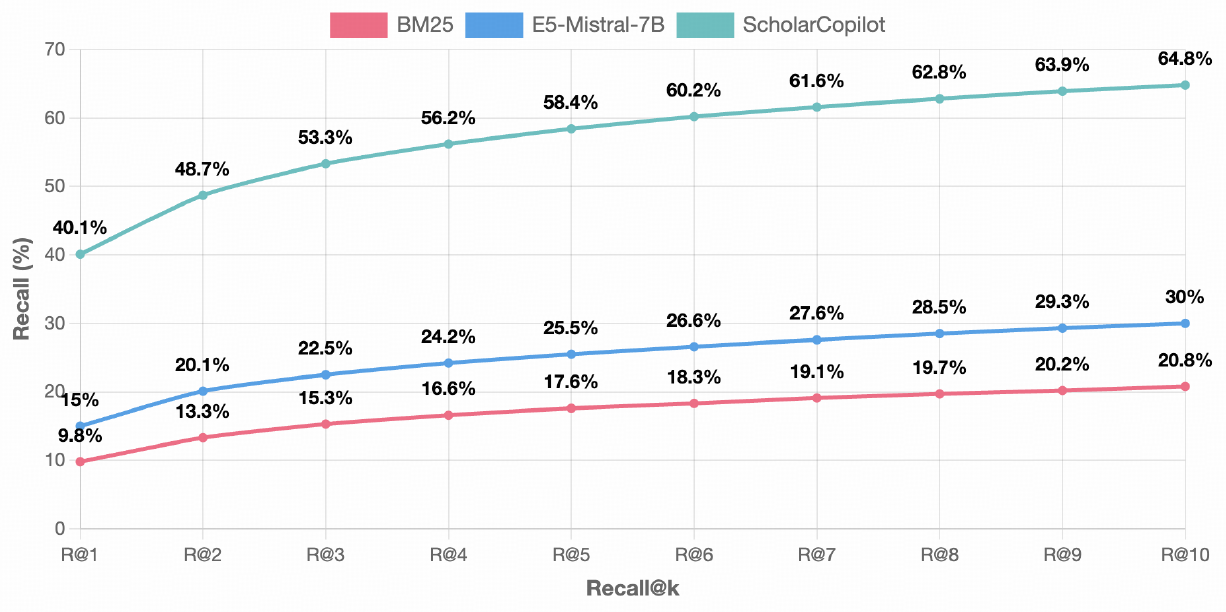}
\caption{Comparison of citation retrieval performance (Recall@k) between ScholarCopilot and baseline retrieval methods (BM25 and E5-Mistral-7B-Instruct). 
}
\label{fig:retrieval_results}
\end{figure}

\subsubsection{Retrieval Performance}

Figure~\ref{fig:retrieval_results} compares ScholarCopilot's citation retrieval performance (Recall@k) with baseline methods. ScholarCopilot achieves a top-1 recall of 40.1\%, significantly outperforming BM25 and E5-Mistral-7B-Instruct. This advantage persists across all recall levels, with ScholarCopilot reaching 64.8\% recall@10, more than doubling E5-Mistral-7B-Instruct and tripling BM25. These results highlight the effectiveness of our unified training approach. Traditional retrieval methods rely on explicitly formulated queries, often failing to capture nuanced citation intents. In contrast, ScholarCopilot directly optimizes retrieval token representations during generation, implicitly encoding citation intent through context-aware queries informed by both local (surrounding text) and global (document-level) information.

\subsubsection{Impact of Reference Content Integration}

We evaluate the impact of providing retrieved reference content to ScholarCopilot during generation by comparing two settings: (1) the standard approach, where the model accesses reference details during generation; and (2) a variant that triggers retrieval but cites papers without seeing their content.

\begin{table}[h]
    \centering
    \small
    \setlength{\tabcolsep}{2pt}
    \begin{tabular}{l|ccccc|c}
    \toprule
    \textbf{Method} & \textbf{Relevance} & \textbf{Coherence} & \textbf{Academic} & \textbf{Completeness} & \textbf{Innovation} & \textbf{Total} \\
    \midrule
    ScholarCopilot & 3.63 & 3.66 & 2.87 & 2.89 & 3.17 & 16.21 \\
    w/o ref. content & 3.60 & 3.25 & 2.58 & 2.91 & 3.19 & 15.53 \\
    \midrule
    $\Delta$ & +0.03 & +0.41 & +0.29 & -0.02 & -0.02 & +0.68 \\
    \bottomrule
    \end{tabular}
    \caption{Impact of reference content integration on generation quality.
    }
    \label{tab:ref_content_ablation}
\end{table}

As shown in Table~\ref{tab:ref_content_ablation}, the two variants perform similarly on Relevance, Completeness, and Innovation. However, differences appear in Coherence (3.66 vs. 3.25) and Academic Rigor (2.87 vs. 2.58), leading to a higher total score for the standard ScholarCopilot (16.21 vs. 15.53). Analysis indicates two reasons. First, access to reference content reduces inaccuracies when describing cited works. Second, reference details provide contextual information that improves coherence, especially during comparisons or transitions between ideas.

Qualitative evaluation shows the variant without reference content tends to cite sources with general statements, whereas the standard approach integrates specific details for clearer connections. For example, when discussing neural networks, the standard model states: "Transformer models leverage self-attention mechanisms to capture long-range dependencies \texttt{cite(vaswani2017attention)}, specifically through a multi-headed approach that projects queries, keys, and values into separate subspaces." In contrast, the variant without reference content produces simpler statements like "Transformer models use self-attention for capturing dependencies \texttt{cite(vaswani2017attention)}."

\section{User Study}

To evaluate the utility of ScholarCopilot in practical academic writing, we conducted a user study with participants from various academic backgrounds. This evaluation assessed both technical performance and user experience.

\subsection{Human Evaluation Design}

We conducted a mixed-method evaluation combining quantitative ratings and qualitative feedback. Participants were 10 students (5 PhD, 4 master's, and 1 undergraduate), averaging 4.2 years of academic writing experience. All participants had prior academic writing experience and were familiar with AI writing assistants such as ChatGPT.

Each participant used ScholarCopilot to draft the introduction and related work sections on at least five topics within their expertise. The evaluation included:\\
\textbf{Quantitative Assessment.} Participants rated ScholarCopilot on 15 metrics using a 5-point Likert scale (1=Poor, 5=Excellent), grouped into Citation Quality (relevance, accuracy, timeliness), User Experience (ease of use, response time, interface clarity, interaction fluidity), and Content Quality (academic rigor, factual accuracy, writing style, logical flow, completeness, topical relevance, innovation, redundancy).\\
\textbf{Comparative Analysis.} Participants compared ScholarCopilot with ChatGPT on citation quality, writing quality, ease of use, time efficiency, and overall usefulness.\\
\textbf{Open-ended Feedback.} Participants commented on ScholarCopilot's strengths, limitations, and suggested improvements.

\subsection{Human Evaluation Results}

\begin{figure}[t]
    \centering
    \begin{subfigure}[b]{0.47\linewidth}
        \centering
        \includegraphics[width=\linewidth]{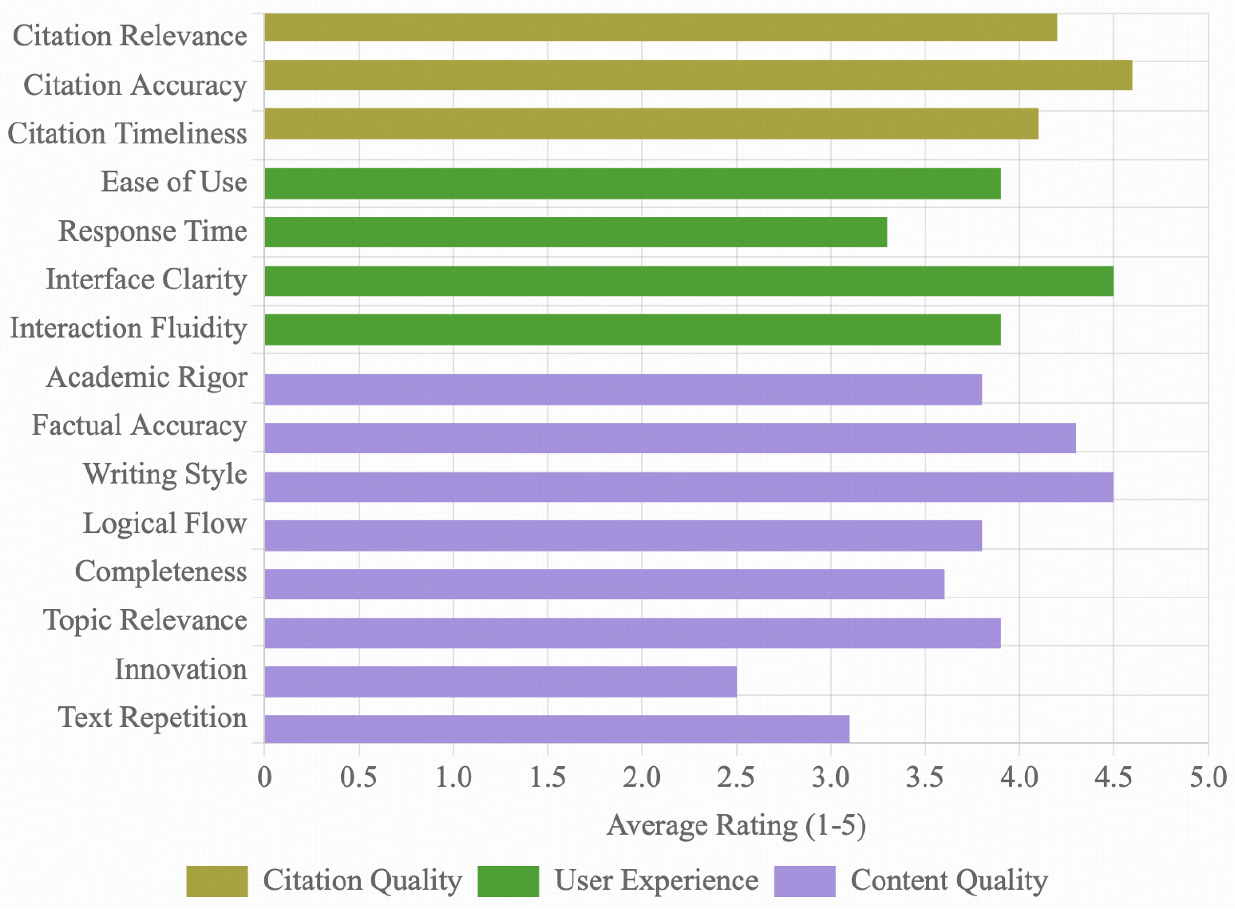}
        \caption{Average ratings for ScholarCopilot across evaluation dimensions: citation quality (yellow), user experience (green), and content quality (purple), from user study (N=10).}
        \label{fig:15_metrics_rating}
    \end{subfigure}
    \hfill
    \begin{subfigure}[b]{0.47\linewidth}
        \centering
        \includegraphics[width=\linewidth]{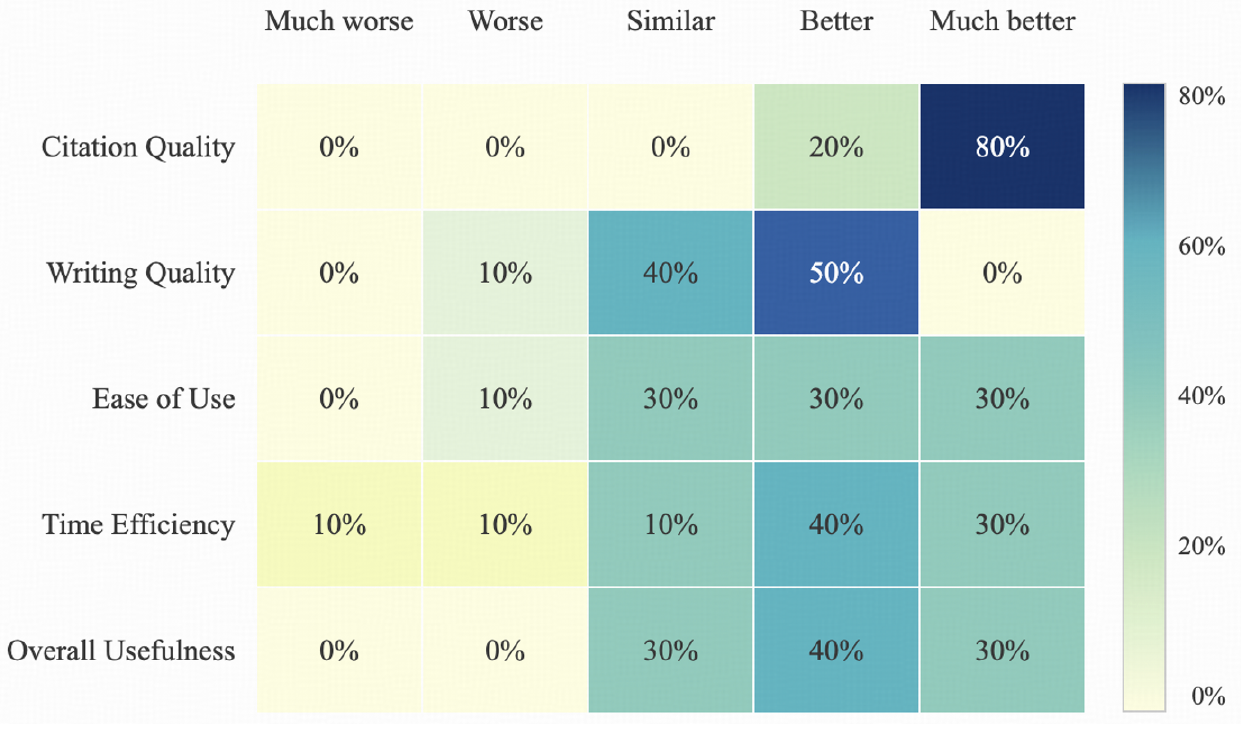}
        \caption{Comparative analysis of ScholarCopilot vs. ChatGPT across five dimensions: Citation Quality, Writing Quality, Ease of Use, Time Efficiency, and Overall Usefulness. Darker blue indicates higher percentages of ratings.}
        \label{fig:comparative_analysis}
    \end{subfigure}
    \caption{Human evaluation of ScholarCopilot and comparative analysis with ChatGPT}
    \label{fig:combined_analysis}
\end{figure}


Figure~\ref{fig:15_metrics_rating} shows the average ratings. ScholarCopilot received the highest scores for citation accuracy (4.6/5), interface clarity (4.5/5), and writing style (4.5/5). Citation quality metrics averaged 4.3/5. User experience metrics averaged 3.9/5, with response time rated lowest (3.3/5).
It is worth noting that the system was deployed on a single 80GB GPU, which led to longer waiting times during peak usage periods.
Due to this resource limitation, different participants experienced significantly varied response times, which explains the inconsistent feedback regarding system responsiveness in the evaluation results.

Content quality metrics showed more variation, with Writing style (4.5/5) and factual accuracy (4.3/5) scoring well, while innovation received the lowest score across all metrics (2.5/5). This suggests that while ScholarCopilot excels at generating academically sound content, it may be less effective at proposing novel ideas or suggesting innovative directions.


\textbf{Comparative Advantage.} Figure~\ref{fig:comparative_analysis} compares ScholarCopilot and ChatGPT.
ScholarCopilot shows a clear advantage in citation quality, with all participants rating ScholarCopilot higher.
For overall usefulness, 70\% rated ScholarCopilot higher.
Writing quality advantage was moderate, with 50\% rating ScholarCopilot higher and 40\% assessing it similar to ChatGPT.\vspace{1ex}\\
\textbf{Qualitative Feedback.} Open-ended responses identified strengths such as integrated citation management (citation search and BibTeX handling), interactive incremental writing style offering greater user control, and improved time efficiency especially for related work sections. Participants also suggested improvements including generating more comprehensive content, reducing system response time for complex retrieval tasks, and enhancing support for generating innovative ideas and research questions.\vspace{1ex}\\
\textbf{Future Use Intention.} Participants' average rating for likelihood of future use was 4.1/5, with 80\% rating this intention as 4 or 5. This suggests ScholarCopilot effectively addresses user needs despite noted limitations.

Participants also provided suggestions, such as integrating with writing platforms like Overleaf, supporting section-wise generation, and allowing predictions at arbitrary cursor positions. These suggestions provide directions for future development.

In summary, the user study confirms ScholarCopilot effectively integrates text generation and citation retrieval, improving user experience in academic writing workflows. The strengths in citation relevance and management indicate advancement over existing tools, while response time and innovation support represent areas for future improvement.

\section{Limitations and Future Work}

Despite promising results, ScholarCopilot currently supports only Introduction and Related Work sections within the computer science domain. Future work will extend the framework to additional paper sections (e.g., methods, experiments) and diverse academic disciplines. Additionally, the user study highlighted limitations in generating innovative insights. Addressing this requires exploring larger models, expanded datasets, and targeted training techniques to enhance creativity. Finally, improvements in user interaction—such as persistent content storage, concise summaries for suggested references, and robust load balancing for consistent multi-user responsiveness—are critical future enhancements.

\section{Conclusion}

We introduced ScholarCopilot, a unified framework integrating dynamic retrieval within the generative process for academic writing. Unlike traditional static retrieval-generation pipelines, ScholarCopilot adaptively retrieves citations based on evolving generation contexts, significantly improving citation accuracy and coherence. Extensive evaluation and user studies demonstrated its effectiveness, particularly in citation relevance, writing efficiency, and overall user experience. Despite current limitations in scope, innovation capability, and interaction design, ScholarCopilot marks a pioneering step toward future advancements in AI-supported academic writing.

\bibliography{colm2025_conference}
\bibliographystyle{colm2025_conference}

\appendix
\section{Appendix}
\subsection{Training Details}
\label{sec:training-details}

We trained our model with the following hyperparameters: maximum context length of 16,384 tokens, learning rate of $1 \times 10^{-5}$, per-device training batch size of 1, and gradient accumulation steps of 4. Training was performed on 4 machines, each equipped with 8 NVIDIA H100 GPUs, resulting in a global batch size of $1 \text{ (per-device batch size)} \times 8 \text{ (GPUs per machine)} \times 4 \text{ (machines)} \times 4 \text{ (gradient accumulation steps)} = 128$.

\lstset{
    basicstyle=\ttfamily\small,      
    columns=fullflexible,            
    frame=single,                    
    breaklines=true,                 
    breakatwhitespace=true,          
    keepspaces=true,                 
    showspaces=false,                
    showstringspaces=false,          
    showtabs=false,                  
    tabsize=4,                       
    rulecolor=\color{gray!50},       
    backgroundcolor=\color{gray!5},  
    language=[LaTeX]TeX,
    literate={\ \ }{{\ }}1           
}
\subsection{Generation Quality Evaluation prompt}
\label{eval_prompt}
\begin{lstlisting}
You are a senior computer science scholar. Please evaluate the AI-generated content using the ground truth as reference.

Evaluate the following five dimensions by comparing the AI-generated content with the ground truth:

[Detailed Evaluation]
1. Content Relevance:
- Key strengths:
- Main gaps:
- Comparison with ground truth:

2. Logical Coherence:
- Key strengths:
- Main gaps:
- Comparison with ground truth:

3. Academic Standards:
- Key strengths:
- Main gaps:
- Comparison with ground truth:

4. Background Completeness:
- Key strengths:
- Main gaps:
- Comparison with ground truth:

5. Innovation Statement:
- Key strengths:
- Main gaps:
- Comparison with ground truth:
[End Evaluation]

[Improvement Suggestions]
1.
2.
3.
[End Suggestions]

Based on your above analysis, provide numerical scores in the following format:
[Scores]
Relevance: <score>/5
Coherence: <score>/5
Academic: <score>/5
Completeness: <score>/5
Innovation: <score>/5
Total: <sum>/25
[End Scores]

Below are the materials for evaluation:

Paper Title:
{title}

Abstract:
{abstract}

Ground Truth Content:
{ground_truth}

AI Generated Content:
{generated_text}

Remember to first provide detailed evaluation, then improvement suggestions, and finally the numerical scores in the exact format specified above."""

\end{lstlisting}

\subsection{Case Study}

\textbf{ScholarCopilot for Academic Writing with Citations Example:}
\begin{lstlisting}
        \title{Noise-Unconditional Diffusion Models for Text-to-Image Generation}
        
        \section{Introduction}
         With the prosperity of deep generative models research, many works emerged focusing on various aspects of generative models, including model architecture design~\cite{prasad2023unraveling, li2023faster, si2023freeu, benny2022dynamic}, training strategies~\cite{xu2024towards, xu2024towards}, and noise scheduling~\cite{chen2023on}. However, there is very limited work discussing whether noise conditioning is a necessary component. Noise conditioning has been an infallible rule widely accepted by researchers in the field of diffusion models research. However, is it really necessary to employ noise conditioning in diffused models? A recent study~\cite{sun2025noise} has challenged this convention. They point out that in some cases diffusion models even perform better without noise conditioning. Inspired by this, we delve deeper into this research direction and explore the feasibility of noise-unconditional diffusion models in the text-to-image generation task.

        In this paper, we propose a novel framework for noise-unconditional diffusion models, named \textbf{Noise-UncondITional Diffusion Models (NUIT)}. Our approach is based on the observation that the noise level is not a necessary input for the denoising process. Previous work~\cite{sun2025noise} has shown that noise conditioning can be removed without significantly compromising performance. Motivated by these findings, we aim to explore the possibility of employing noise-unconditional diffusion models in text-to-image generation tasks. Compared to general image generation tasks, text-to-image generation tasks are more challenging and can better reflect the model's generalization ability. To this end, we focus on the text-to-image generation task in this paper.
        
        Our approach is based on the idea that the noise level can be inferred from the model's internal states, rather than being explicitly provided as input. We propose a novel architecture that eliminates the need for explicit noise level inputs. This architecture is based on the idea that the noise level can be inferred from the model's internal states, rather than being explicitly provided as input. Specifically, we propose to use a self-attention mechanism to capture the global information of the image, and then use this information to infer the noise level. This approach allows us to remove the need for explicit noise level inputs, while still achieving competitive performance on standard benchmarks. Our approach demonstrates that noise-unconditional diffusion models can be a viable alternative to traditional noise-conditioned diffusion models in text-to-image generation tasks. Our contributions can be summarized as follows:
        
        \begin{itemize}
            \item We propose a novel framework for noise-unconditional diffusion models, which eliminates the need for explicit noise level inputs.
            \item We demonstrate that noise-unconditional diffusion models can achieve competitive performance on standard benchmarks, achieving state-of-the-art FID and CLIP score.
            \item We provide an in-depth analysis of the noise-unconditional diffusion models, and discuss the potential applications of this approach in various domains.
        \end{itemize}
        
        We believe that our work will inspire more research in this direction, and will contribute to the development of more efficient and effective generative models.


Bibliography:

@article{prasad2023unraveling,
title={Unraveling the Temporal Dynamics of the Unet in Diffusion Models},
author={Vidya Prasad, Chen Zhu-Tian, Anna Vilanova, Hanspeter Pfister, Nicola
Pezzotti, Hendrik Strobelt},
journal={arXiv preprint arXiv:2312.14965},
year={2023},
archivePrefix={arXiv},
eprint={2312.14965},
primaryClass={cs.CV cs.LG}
}

@article{li2023faster,
title={Faster Diffusion: Rethinking the Role of the Encoder for Diffusion Model
Inference},
author={Senmao Li, Taihang Hu, Joost van de Weijer, Fahad Shahbaz Khan, Tao
Liu, Linxuan Li, Shiqi Yang, Yaxing Wang, Ming-Ming Cheng and Jian Yang},
journal={arXiv preprint arXiv:2312.09608},
year={2023},
archivePrefix={arXiv},
eprint={2312.09608},
primaryClass={cs.CV}
}

@article{si2023freeu:,
title={FreeU: Free Lunch in Diffusion U-Net},
author={Chenyang Si, Ziqi Huang, Yuming Jiang, Ziwei Liu},
journal={arXiv preprint arXiv:2309.11497},
year={2023},
archivePrefix={arXiv},
eprint={2309.11497},
primaryClass={cs.CV}
}

@article{benny2022dynamic,
title={Dynamic Dual-Output Diffusion Models},
author={Yaniv Benny, Lior Wolf},
journal={arXiv preprint arXiv:2203.04304},
year={2022},
archivePrefix={arXiv},
eprint={2203.04304},
primaryClass={cs.CV eess.IV}
}

@article{xu2024towards,
title={Towards Faster Training of Diffusion Models: An Inspiration of A
Consistency Phenomenon},
author={Tianshuo Xu, Peng Mi, Ruilin Wang, Yingcong Chen},
journal={arXiv preprint arXiv:2404.07946},
year={2024},
archivePrefix={arXiv},
eprint={2404.07946},
primaryClass={cs.LG cs.AI}
}

@article{chen2023on,
title={On the Importance of Noise Scheduling for Diffusion Models},
author={Ting Chen},
journal={arXiv preprint arXiv:2301.10972},
year={2023},
archivePrefix={arXiv},
eprint={2301.10972},
primaryClass={cs.CV cs.GR cs.LG cs.MM}
}

@article{sun2025noise,
title={Is Noise Conditioning Necessary for Denoising Generative Models?},
author={Sun, Qiao and Jiang, Zhicheng and Zhao, Hanhong and He, Kaiming},
journal={arXiv preprint arXiv:2502.13129},
year={2025}
}
\end{lstlisting}

\textbf{Analysis:}

The ScholarCopilot-generated introduction demonstrates good coherence, logical structure, and a clear presentation of motivation, technical concepts, and contributions. It exhibits fluent academic English and accurate citation formatting, reflecting effective training on academic writing conventions. However, the text occasionally contains redundant phrases, such as repeated descriptions about inferring noise levels from internal model states. Additionally, briefly contextualizing and contrasting your method against related works would help further emphasize the originality and significance of your contributions.

\textbf{GPT-4o for Academic Writing with Citations Example:}
\begin{lstlisting}
        \title{Noise-Unconditional Diffusion Models for Text-to-Image Generation}
        
        \begin{abstract}
        Diffusion models have emerged as powerful tools for generative tasks, particularly in the domain of image synthesis. While conditional diffusion models have achieved remarkable success in converting text prompts into high-quality images, the reliance on explicit conditioning signals often limits their generalizability. In this paper, we propose a novel paradigm: Noise-Unconditional Diffusion Models (NUDMs), which decouple the text encoding process from the noise generation process, enabling enhanced flexibility and robustness in text-to-image generation tasks.
        \end{abstract}
        
        \section{Introduction}
        
        Text-to-image generation has become a cornerstone of modern AI research, enabling systems to translate natural language descriptions into visually coherent and semantically aligned images. Recent advancements in diffusion models have demonstrated their potential in generating photorealistic and diverse images, making them a competitive alternative to adversarial models such as GANs \cite{goodfellow2014gan}. Conditional diffusion models, such as those used in DALLE 2 \cite{ramesh2022dalle2} and Imagen \cite{saharia2022imagen}, leverage text embeddings to guide the denoising process, ensuring that the generated images align with the input text prompt.
        
        Despite their successes, conditional diffusion models face challenges related to overfitting to conditioning signals \cite{ho2020denoising} and the computational overhead of processing complex text-image alignments. Additionally, the reliance on explicit conditioning introduces a bottleneck when dealing with noisy or ambiguous text inputs. To address these issues, we introduce the concept of Noise-Unconditional Diffusion Models (NUDMs), which operate independently of explicit text conditioning during the noise generation phase. Instead, text embeddings are integrated into the model through an auxiliary alignment mechanism, allowing for more robust and flexible generation.
        
        Our contributions are threefold. First, we propose a noise-unconditional framework for diffusion-based text-to-image generation, which separates the noise generation process from textual conditioning. Second, we introduce a novel alignment module that bridges the gap between text and image embeddings, enhancing semantic coherence. Finally, we demonstrate through extensive experiments that NUDMs outperform existing conditional diffusion models in terms of both image quality and text-image alignment, especially under noisy or ambiguous text prompts.
        
        The rest of this paper is organized as follows. Section \ref{sec:related_work} reviews related work in diffusion models and text-to-image synthesis. Section \ref{sec:methodology} details the proposed Noise-Unconditional Diffusion Model framework. Section \ref{sec:experiments} presents experimental results and evaluations. Finally, Section \ref{sec:conclusion} concludes the paper with insights and future directions.

Bibliography:

@inproceedings{goodfellow2014gan,
  title={Generative adversarial nets},
  author={Goodfellow, Ian and Pouget-Abadie, Jean and Mirza, Mehdi and Xu, Bing and Warde-Farley, David and Ozair, Sherjil and Courville, Aaron and Bengio, Yoshua},
  booktitle={Advances in neural information processing systems},
  pages={2672--2680},
  year={2014}
}

@article{ramesh2022dalle2,
  title={Hierarchical Text-Conditional Image Generation with CLIP Latents},
  author={Ramesh, Aditya and Dhariwal, Prafulla and Nichol, Alex and Chu, Casey and Chen, Mark},
  journal={arXiv preprint arXiv:2204.06125},
  year={2022}
}

@article{saharia2022imagen,
  title={Imagen: Text-to-image diffusion models with large pre-trained language models},
  author={Saharia, Chitwan and Chan, William and Saxena, Saurabh and Li, Lala and Whang, Jay and Denton, Emily and Ghasemipour, Seyed Kamyar Seyed and Ayan, Burcu Karagol and Mahdavi, Shuyang and Lopes, Raphael and others},
  journal={arXiv preprint arXiv:2205.11487},
  year={2022}
}

@article{ho2020denoising,
  title={Denoising diffusion probabilistic models},
  author={Ho, Jonathan and Jain, Ajay and Abbeel, Pieter},
  journal={Advances in Neural Information Processing Systems},
  volume={33},
  pages={6840--6851},
  year={2020}
}

\end{lstlisting}

The content generated by GPT-4o is also highly fluent, demonstrating strong coherence in structure and effectively conveying the rationale, technical details, and key contributions in a distinct style. However, it similarly suffers from insufficient depth and inadequate background context. Furthermore, it lacks comprehensive referencing to relevant literature, citing only four highly influential papers (each with over 5000 citations). Notably, it also exhibits a hallucination regarding the Imagen paper, incorrectly stating its title; the correct title should be ``Photorealistic Text-to-Image Diffusion Models with Deep Language Understanding.''

\textbf{Qwen-2.5-72B-re for Academic Writing with Citations Example: }

\begin{lstlisting}
        \title{Noise-Unconditional Diffusion Models for Text-to-Image Generation}
        
        \section{Introduction}
        
        In recent years, deep generative models have become popular, leading to numerous research efforts focusing on improving various aspects, such as architecture of models~\cite{goodfellow2014generative, kingma2013auto}, training methods~\cite{he2016deep}, and strategies for scheduling noise~\cite{vaswani2017attention}. Among these aspects, noise conditioning has typically been accepted by the research community as a necessary and fundamental component in diffusion models. Nonetheless, there is limited exploration of whether this assumption holds true universally. A recent preliminary study by Sun et al.~\cite{sun2025noise} suggests that noise conditioning may not always be necessary, and in some scenarios, its absence could potentially lead to improved results.
        
        Motivated by these initial findings, this paper introduces a noise-unconditional diffusion model, called \textbf{NU-DM}, specifically designed for text-to-image generation. Unlike previous studies, we do not explicitly provide the noise level information as input to the diffusion model. Instead, our model internally infers the noise level implicitly from the learned representation of the image during training. To achieve this, we propose using a convolutional neural network (CNN) architecture to extract global image features, which indirectly encode noise information. By removing explicit noise conditioning, the NU-DM model simplifies the diffusion process and potentially reduces model complexity.
        
        We evaluate our NU-DM model on several standard benchmarks in text-to-image generation, demonstrating performance that is competitive with previous state-of-the-art diffusion models~\cite{krizhevsky2012imagenet}. In our experiments, we measured commonly used evaluation metrics such as the Frechet Inception Distance (FID) and CLIP-based alignment scores~\cite{lecun1998gradient}, showing comparable results to current methods.
        
        Our contribution can be summarized as follows:
        
        \begin{itemize}
        \item We introduce a new diffusion model (NU-DM) that operates without explicit noise conditioning, simplifying the model structure.
        \item We conduct experiments on text-to-image datasets, achieving competitive performance compared to previous works.
        \item We carry out analysis on NU-DM, discussing the effectiveness and limitations of removing explicit noise conditioning.
        \end{itemize}
        
        We hope that our study will encourage further investigation into noise conditioning and inspire new diffusion model designs.
\end{lstlisting}

\textbf{Analysis:}

Qwen-2.5-72B-re generates reasonably coherent introduction sections, capturing the general structure and flow expected in academic papers. However, the content often lacks sufficient depth, and the technical descriptions tend to be superficial and repetitive. Moreover, the citation accuracy is notably poor, with several incorrect and irrelevant references included. For example, citations intended to support diffusion modeling techniques incorrectly reference well-known papers on unrelated methods, such as GANs and Transformers. These issues highlight significant room for improvement in both the accuracy of citations and the richness of technical content.

\textbf{Qwen-2.5-7B-re for Academic Writing with Citations Example:}
\begin{lstlisting}

        \title{Noise-Unconditional Diffusion Models for Text-to-Image Generation}
        
        \section{Introduction}
        
        Recently, diffusion models have become popular in computer vision because they are good at generating images. Diffusion models usually require noise conditioning, which helps them produce realistic images~\cite{he2016deep, kingma2013auto}. However, some researchers found noise conditioning may not always be necessary, though this question is still not fully answered. For example, Vaswani et al.~\cite{vaswani2017attention} showed results without noise, but their experiments were not sufficient.
        
        Motivated by this unclear situation, we propose a new diffusion model called NU-Diffusion, meaning ``Noise-Unconditional Diffusion.'' Our NU-Diffusion model generates images from text inputs without explicitly using noise conditioning. Specifically, we use convolutional neural networks (CNNs)~\cite{lecun1998gradient} to implicitly infer noise information from the input data. We chose CNNs because they have been successfully used in many image-related tasks and generally perform well.
        
        We conduct experiments on text-to-image datasets to test our NU-Diffusion model. Experimental results show that our model can generate high-quality images effectively. The generated images look good and are visually appealing. Additionally, we use common metrics like accuracy and FID scores to evaluate our results, achieving good performance compared to other existing approaches~\cite{krizhevsky2012imagenet}. We hope our study will inspire more researchers to investigate noise conditioning further.

\end{lstlisting}

\textbf{Analysis:}

Qwen-2.5-7B-re-generated introduction demonstrates noticeable weaknesses in citation accuracy, logical coherence, and technical depth. Citations such as he2016deep, vaswani2017attention, krizhevsky2012imagenet are incorrectly used, indicating misunderstanding of relevant literature. Additionally, informal expressions (``images look good and are visually appealing'') and inappropriate evaluation metrics (``accuracy'') further undermine its academic rigor. Overall, the baseline introduction clearly shows substantial room for improvement in technical correctness and scholarly style.

\subsection{Human Study Questionnaire Details}

In the following part, we provide the specific details of the Human Study Questionnaire

\begin{figure}[ht]
\centering
\includegraphics[width=\linewidth]{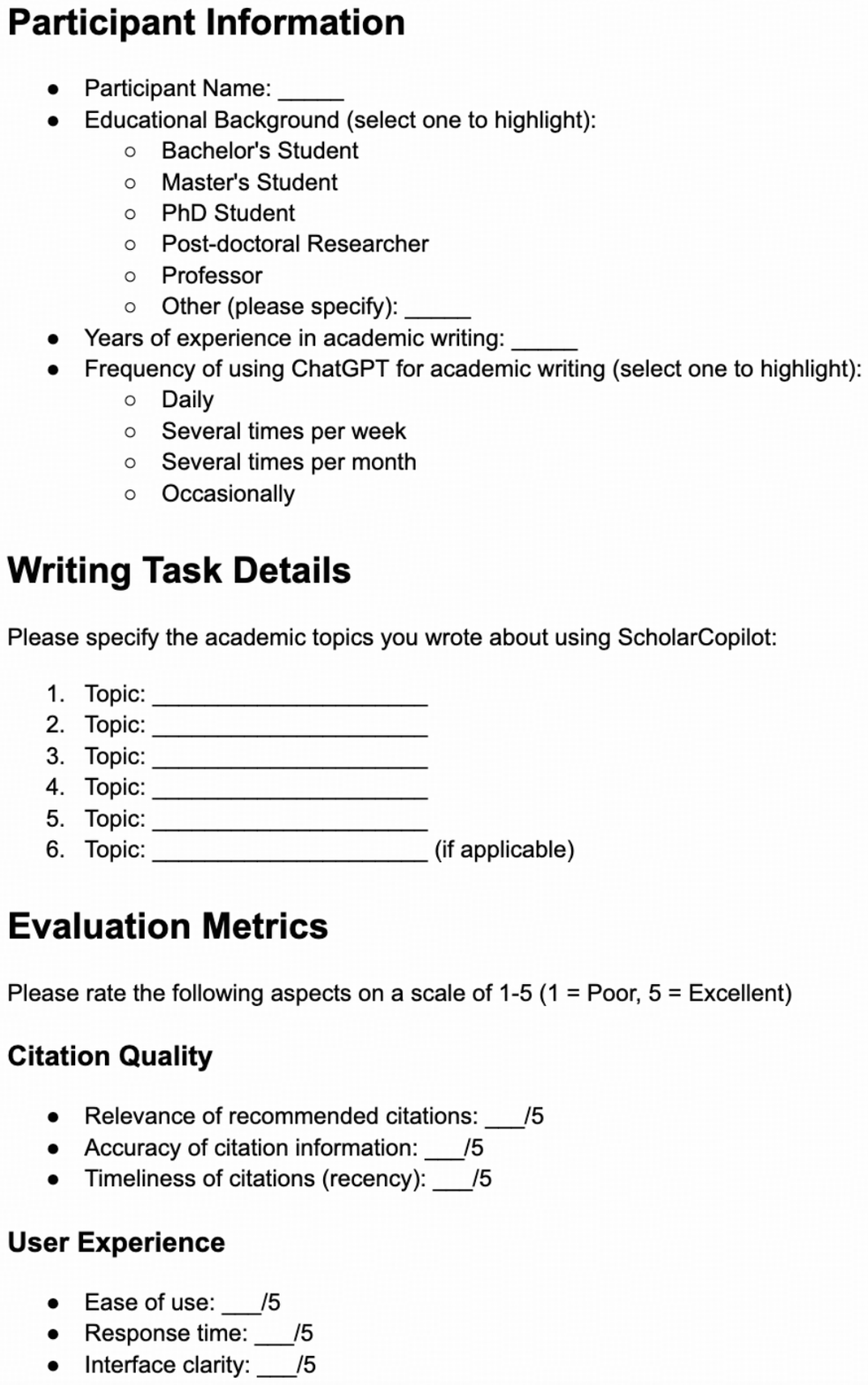}
\caption{Human Study Questionnaire Page 1}
\end{figure}

\begin{figure}[ht]
\centering
\includegraphics[width=\linewidth]{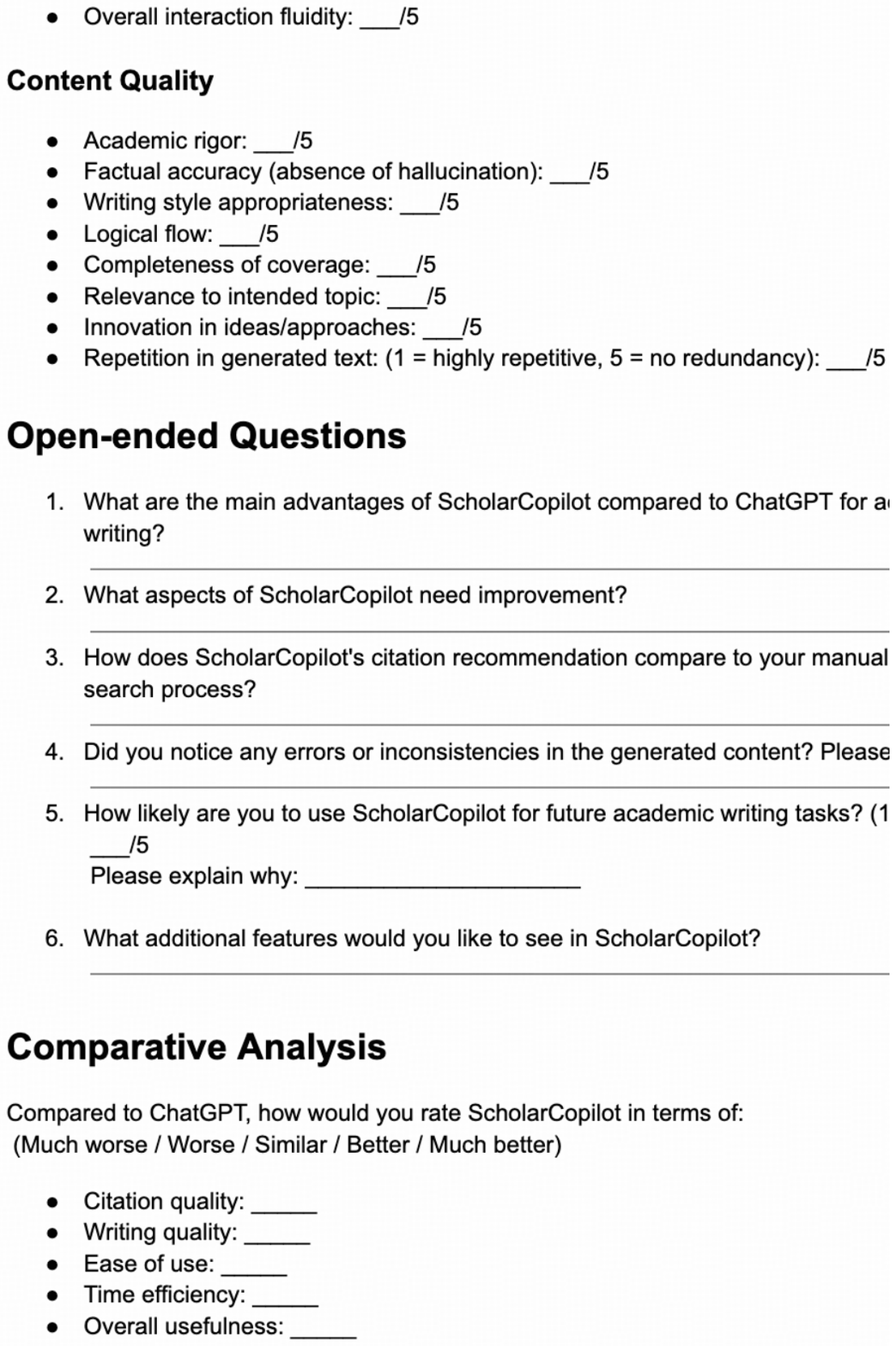}
\caption{Human Study Questionnaire Page 2}
\end{figure}

\begin{figure}[ht]
\centering
\includegraphics[width=\linewidth]{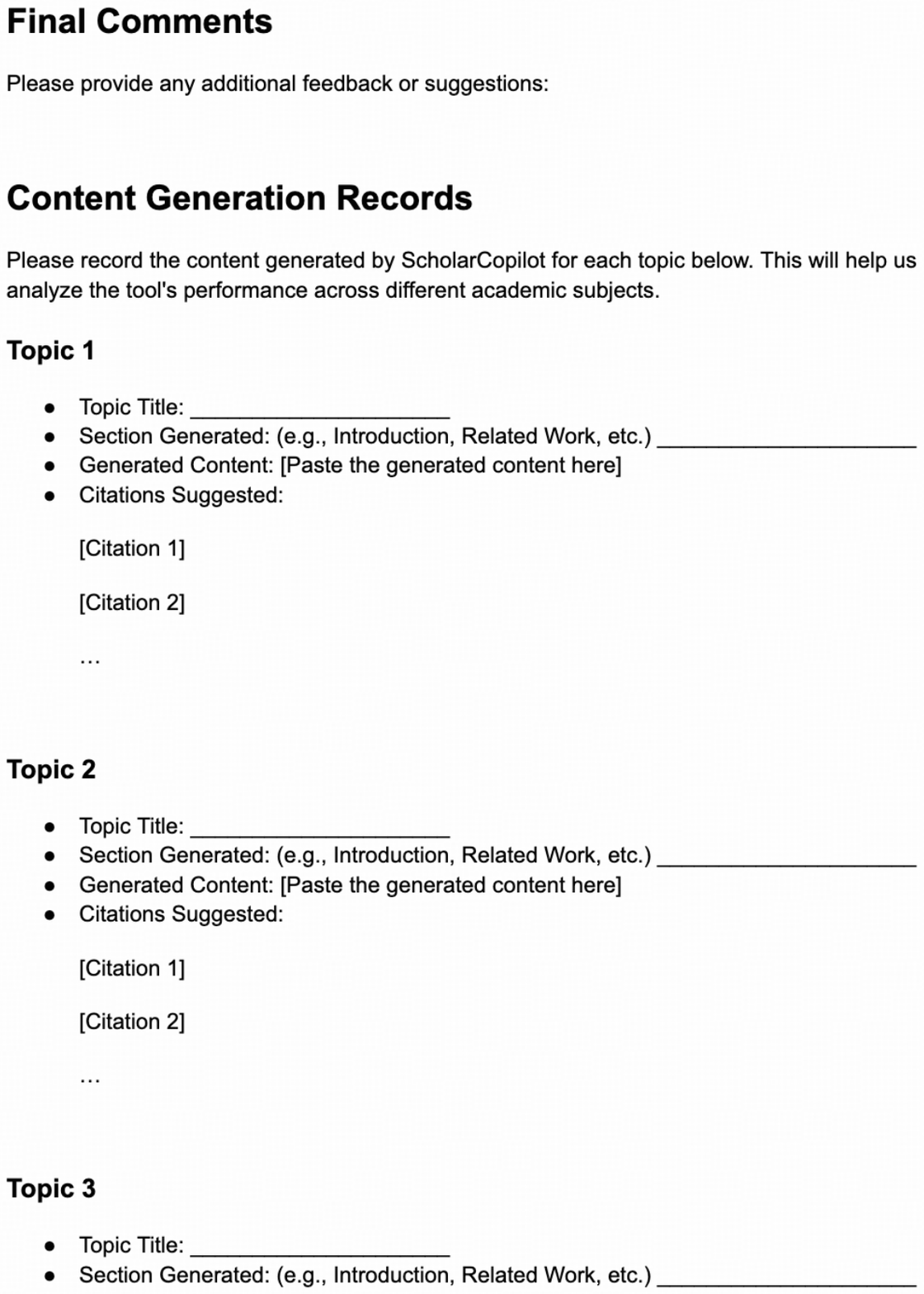}
\caption{Human Study Questionnaire Page 3}
\end{figure}

\begin{figure}[ht]
\centering
\includegraphics[width=\linewidth]{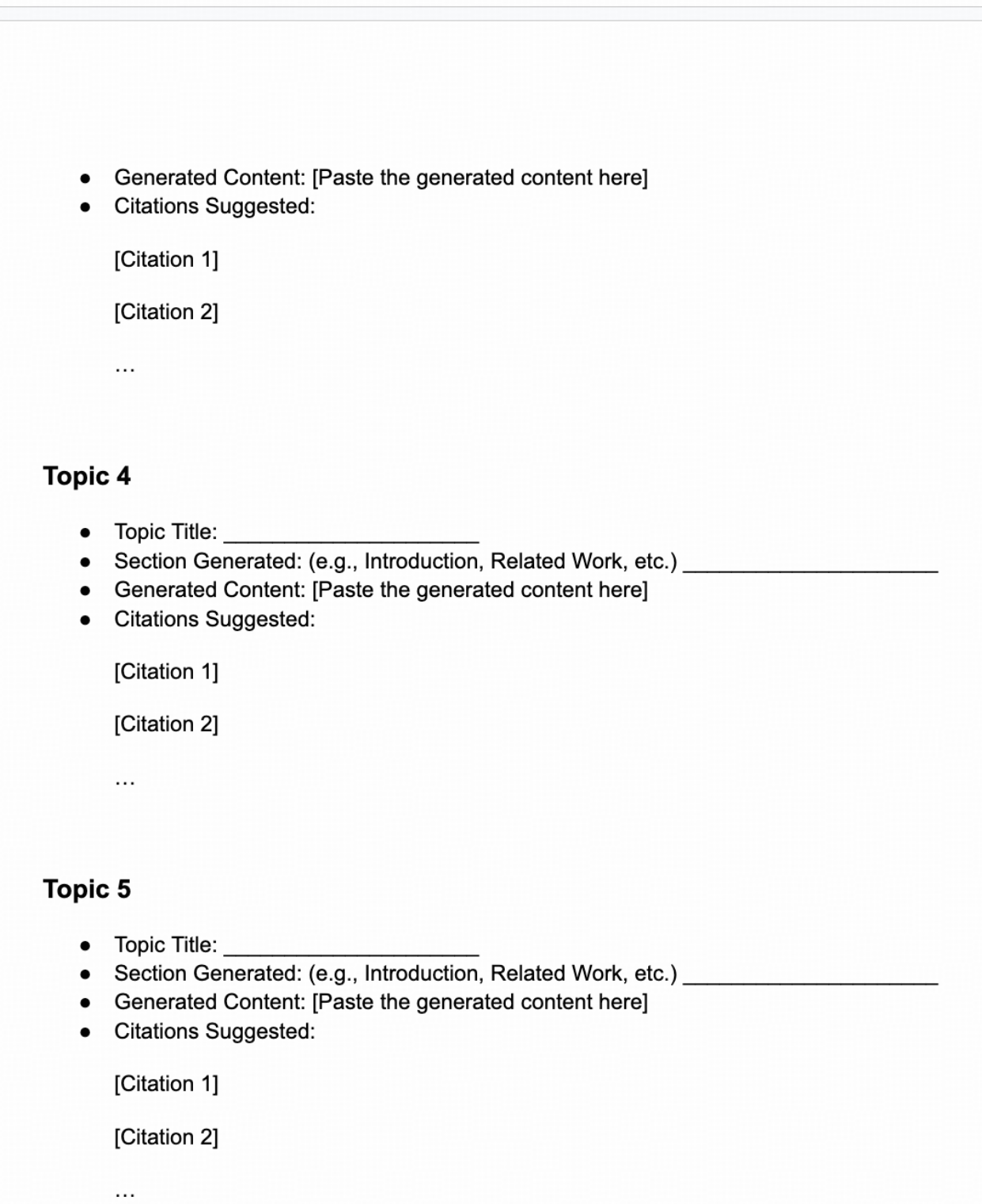}
\caption{Human Study Questionnaire Page 4}
\end{figure}

\end{document}